\documentclass[letterpaper]{article}

\usepackage{aaai2026}
\usepackage{times}
\usepackage{helvet}
\usepackage{courier}
\usepackage[hyphens]{url}
\usepackage{graphicx}
\usepackage{natbib}
\usepackage{caption}
\usepackage{booktabs}
\usepackage{array}
\usepackage{makecell}
\usepackage{threeparttable}
\usepackage{amsmath}
\usepackage{indentfirst}
\usepackage{microtype}
\usepackage{amsmath}
\usepackage{amssymb}
\usepackage{amsfonts}
\usepackage{booktabs}
\usepackage{threeparttable}
\usepackage{graphicx}

\usepackage{multirow}
\usepackage{xcolor}
\usepackage{colortbl}

\newcommand{\rankone}[1]{\textbf{#1}}
\newcommand{\ranktwo}[1]{\underline{#1}}
\newcommand{\rankthree}[1]{#1}
\newcommand{\rankfour}[1]{#1}
\newcommand{\rankfive}[1]{#1}

\newcommand{\MainComparisonRows}{%

    \midrule
    \multicolumn{13}{@{}l}{\textbf{A. Fully-supervised}} \\

    PraNet~\cite{fan2020pranet}
    & 90.16 & 13.48
    & 90.53 & 4.01
    & 71.70 & 42.30
    & 87.68 & 9.87
    & 63.64 & 121.12
    & 80.74 & 38.15 \\

    nnU-Net~\cite{isensee2021nnunet}
    & 81.91 & 23.60
    & 79.89 & 11.20
    & 60.97 & 56.45
    & 78.61 & 22.52
    & 50.66 & 159.88
    & 70.41 & 54.73 \\

    HarDNet~\cite{huang2021hardnetmseg}
    & 89.80 & 13.43
    & 91.15 & 4.81
    & 71.82 & 36.22
    & 88.55 & 7.03
    & 70.20 & 78.90
    & 82.30 & 28.08 \\

    Polyp-PVT~\cite{dong2023polyppvt}
    & 91.98 & 9.81
    & 94.03 & 2.36
    & 81.35 & 18.72
    & 90.39 & 6.21
    & 79.01 & 41.71
    & 87.35 & 15.76 \\

    MDPNet~\cite{kamara2025mdpnet}
    & 86.63 & 18.79
    & 85.56 & 15.49
    & 62.36 & 18.45
    & 84.23 & 10.58
    & 52.47 & 67.71
    & 74.25 & 26.20 \\

    \midrule
    \multicolumn{13}{@{}l}{\textbf{B. Foundation Model}} \\

    FastSAM+Oracle Box~\cite{zhao2023fastsam}
    & 51.52 & 107.33
    & 39.68 & 76.00
    & 27.87 & 159.08
    & 37.36 & 128.04
    & 34.39 & 296.08
    & 38.16 & 153.31 \\

    SAM-Med2D~\cite{cheng2023sammed2d}
    & 92.17 & 10.11
    & 86.95 & 6.38
    & 90.05 & 5.42
    & 91.50 & 4.13
    & 87.34 & 11.18
    & 89.60 & 7.44 \\

    MedSAM2~\cite{zhu2024medsam2}
    & 91.34 & 10.94
    & 81.87 & 6.54
    & 90.95 & 4.75
    & 91.68 & 3.89
    & 91.54 & 6.62
    & 89.48 & 6.55 \\

    ScribblePrompt~\cite{wong2024scribbleprompt}
    & 74.53 & 25.67
    & 74.48 & 8.77
    & 75.45 & 14.75
    & 80.28 & 8.16
    & 67.70 & 25.63
    & 74.49 & 16.60 \\

    SAMed-2~\cite{yan2025samed2}
    & 79.89 & 24.25
    & 81.56 & 7.37
    & 82.28 & 10.78
    & 88.20 & 5.12
    & 81.34 & 18.19
    & 82.65 & 13.14 \\

    SAM3 No Prompt~\cite{carion2025sam3}
    & 26.98 & 170.13
    & 15.19 & 95.01
    & 13.20 & 160.95
    & 6.68 & 182.68
    & 8.02 & 440.14
    & 14.01 & 209.78 \\

    SAM3+Point Prompt~\cite{carion2025sam3}
    & 81.15 & 24.49
    & 88.32 & 7.45
    & 81.52 & 16.01
    & 84.41 & 12.21
    & 80.63 & 33.74
    & 83.21 & 18.78 \\

    SAM3+Box Prompt~\cite{carion2025sam3}
    & 85.80 & 13.94
    & 92.14 & 2.93
    & 92.26 & 4.20
    & 93.27 & 3.08
    & 93.98 & 6.29
    & 91.49 & 6.09 \\

    SAM3+Gaze Prompt~\cite{carion2025sam3}
    & 81.18 & 26.50
    & 83.47 & 11.07
    & 81.09 & 18.48
    & 85.85 & 11.03
    & 77.14 & 53.38
    & 81.75 & 24.09 \\

    \midrule
    \multicolumn{13}{@{}l}{\textbf{C. Weakly-supervised}} \\

    BoxSup~\cite{dai2015boxsup}\tnote{a,b}
    & 24.57 & 191.11
    & 20.07 & 67.66
    & 12.54 & 146.96
    & 6.66 & 163.12
    & 8.64 & 318.69
    & 14.50 & 177.51 \\

    WeSAM~\cite{zhang2024wesam}
    & 71.97 & 31.29
    & 74.05 & 11.15
    & \rankfive{76.20} & \rankfive{14.67}
    & 83.76 & 8.39
    & \rankfour{79.66} & \rankone{28.28}
    & \rankfive{77.13} & \rankfive{18.75} \\

    WeakMedSAM~\cite{wang2025weakmedsam}\tnote{b}
    & 22.84 & 105.37
    & 12.85 & 83.06
    & 12.01 & 120.68
    & 11.52 & 122.18
    & 9.85 & 243.87
    & 13.81 & 135.03 \\

    WeakPolyp-SAM~\cite{zhao2025weakpolypsam}
    & \rankfive{84.42} & 19.44
    & \rankfive{84.18} & \rankfive{8.29}
    & 65.44 & 32.46
    & \rankfive{87.20} & \rankfive{8.15}
    & 63.42 & 59.04
    & 76.93 & 25.48 \\

    \midrule
    \multicolumn{13}{@{}l}{\textbf{D. Gaze-supervised}} \\

    GazeMedSeg~\cite{zhong2024gazemedseg}
    & 75.28 & 27.57
    & 65.55 & 27.05
    & 54.40 & 60.04
    & 66.80 & 22.99
    & 41.56 & 159.76
    & 60.72 & 59.48 \\

    FGI~\cite{chen2025fgi}
    & 74.07 & 24.85
    & 26.71 & 42.31
    & 14.93 & 60.68
    & 26.40 & 26.05
    & 28.13 & 114.91
    & 34.05 & 53.79 \\

    GNAN~\cite{wu2025gnan}
    & 74.18 & 37.22
    & 59.74 & 28.54
    & 51.80 & 124.23
    & 67.58 & 20.14
    & 42.36 & 358.72
    & 59.13 & 113.77 \\

    GiTNet~\cite{wu2026gitnet}
    & 77.34 & 30.34
    & 60.68 & 30.14
    & 50.17 & 118.82
    & 59.72 & 42.85
    & 41.62 & 355.38
    & 57.91 & 115.51 \\

    GAM~\cite{ge2026gam}
    & 81.68 & \rankfive{17.74}
    & 78.79 & 8.85
    & 73.20 & 17.60
    & 81.25 & 10.02
    & 66.54 & \rankfive{49.91}
    & 76.29 & 20.82 \\

    \midrule
    \multicolumn{13}{@{}l}{\textbf{E. Ours}} \\

    GEG-SAM (SAM-Base)
    & \rankthree{88.70} & \rankthree{12.47}
    & \rankfour{85.95} & \rankone{6.41}
    & \rankfour{83.29} & \rankfour{13.05}
    & \rankfour{89.55} & \rankone{5.15}
    & \rankfive{78.63} & \rankthree{46.33}
    & \rankfour{85.23} & \ranktwo{16.68} \\

    GEG-SAM (SAM-Large)
    & \ranktwo{89.12} & \rankone{11.27}
    & \rankthree{86.20} & \rankthree{6.89}
    & \ranktwo{84.68} & \rankone{11.95}
    & \rankthree{89.97} & \ranktwo{5.35}
    & \rankthree{79.98} & \ranktwo{44.71}
    & \ranktwo{85.99} & \rankone{16.03} \\

    GEG-SAM (SAM-Huge)
    & \rankfour{87.77} & \rankfour{13.58}
    & \rankone{87.21} & \ranktwo{6.51}
    & \rankthree{84.03} & \rankthree{12.70}
    & \ranktwo{90.04} & \rankthree{5.65}
    & \rankone{80.66} & \rankfour{47.10}
    & \rankthree{85.94} & \rankthree{17.11} \\

    \textbf{GEG-SAM (SAM3)}
    & \rankone{89.51} & \ranktwo{12.29}
    & \ranktwo{86.86} & \rankfour{7.56}
    & \rankone{85.26} & \ranktwo{12.09}
    & \rankone{90.35} & \rankfour{5.92}
    & \ranktwo{80.16} & 54.39
    & \rankone{86.43} & \rankfour{18.45} \\
}

\newcommand{\MainResultsTable}{%
\begin{table*}[t]
    \centering
    \begin{threeparttable}

    \small
    \setlength{\tabcolsep}{1.00pt}
    \renewcommand{\arraystretch}{1.00}

    \caption{Comparison on five public polyp segmentation datasets. Dice (\%) and
    ASSD (pixels) are reported, with Mean denoting the average over all datasets.
    The best and second-best results among Sections C--E are marked in bold and
    underlined.}
    \label{tab:main_results}

    \begingroup
    \setlength{\arrayrulewidth}{0.2pt}

    \begin{tabular}{@{}l|*{10}{c}|*{2}{c}@{}}
        \toprule

        \multirow{2}{*}{\textbf{Method}}
        & \multicolumn{2}{|c}{\textbf{Kvasir-SEG}}
        & \multicolumn{2}{c}{\textbf{ClinicDB}}
        & \multicolumn{2}{c}{\textbf{ColonDB}}
        & \multicolumn{2}{c}{\textbf{CVC-300}}
        & \multicolumn{2}{c}{\textbf{ETIS}}
        & \multicolumn{2}{|c}{\textbf{Mean}} \\

        \cmidrule(lr){2-3}
        \cmidrule(lr){4-5}
        \cmidrule(lr){6-7}
        \cmidrule(lr){8-9}
        \cmidrule(lr){10-11}
        \cmidrule(lr){12-13}

        & \textbf{Dice} $\uparrow$ & \textbf{ASSD} $\downarrow$
        & \textbf{Dice} $\uparrow$ & \textbf{ASSD} $\downarrow$
        & \textbf{Dice} $\uparrow$ & \textbf{ASSD} $\downarrow$
        & \textbf{Dice} $\uparrow$ & \textbf{ASSD} $\downarrow$
        & \textbf{Dice} $\uparrow$ & \textbf{ASSD} $\downarrow$
        & \textbf{Dice} $\uparrow$ & \textbf{ASSD} $\downarrow$ \\

        \MainComparisonRows

        \bottomrule
    \end{tabular}

    \endgroup

    \begin{tablenotes}[flushleft]
        \small
        \item[a] BoxSup is reproduced using a third-party PyTorch implementation.
        \item[b] Category-level supervision is less effective in our single-category setting.
    \end{tablenotes}

    \end{threeparttable}
\end{table*}
}

\providecommand{\cmark}{\checkmark}

\newcommand{\ComponentAblationTable}{%
\begin{table}[t]
    \centering
    \small
    \setlength{\tabcolsep}{4.2pt}
    \renewcommand{\arraystretch}{0.98}
    \caption{Ablation on main components. Cross-Domain denotes the average over
    CVC-ColonDB, CVC-300, and ETIS. All entries report Dice (\%).}
    \label{tab:component_ablation}
    \begin{tabular}{@{}ccc|cccc@{}}
        \toprule
        \textbf{GDSE}
        & \textbf{GSTC}
        & \textbf{GFI}
        & \textbf{\shortstack{Kvasir-\\SEG}}
        & \textbf{\shortstack{CVC-\\ClinicDB}}
        & \textbf{\shortstack{Cross-\\Domain}}
        & \textbf{Mean} \\
        \midrule
          &   &   & 80.81 & 83.42 & 78.89 & 80.18 \\
        \cmark &   &   & 86.04 & 84.15 & 83.25 & 83.99 \\
          & \cmark &   & 87.66 & 86.16 & 84.54 & 85.49 \\
        \cmark & \cmark &   &
        \underline{88.24} &
        \textbf{87.61} &
        \underline{84.92} &
        \underline{86.12} \\
        \cmark & \cmark & \cmark &
        \textbf{89.81} &
        \underline{86.69} &
        \textbf{85.50} &
        \textbf{86.60} \\
        \bottomrule
    \end{tabular}
\end{table}
}

\title{GEG-SAM: Gaze Exploration-Guided Segment Anything Model for Weakly Supervised Polyp Segmentation}

\author{
    Ao Wang\textsuperscript{\rm 1},
    Lei Zhang\textsuperscript{\rm 2},
    Lituan Wang\textsuperscript{\rm 2}\protect\thanks{
        Lituan Wang and Zhenwei Zhang are co-corresponding authors.
    },
    Zhenwei Zhang\textsuperscript{\rm 1,*},
    Xiangbao Gong\textsuperscript{\rm 1},\\
    Zhiyi Zhang\textsuperscript{\rm 1},
    Shunyao Wang\textsuperscript{\rm 3},
    Xianglei Yuan\textsuperscript{\rm 3},
    Zizhou Wang\textsuperscript{\rm 4}
}

\affiliations{
    \textsuperscript{\rm 1}School of Computer Science, Sichuan University, China\\
    \textsuperscript{\rm 2}School of Artificial Intelligence, Sichuan University, China\\
    \textsuperscript{\rm 3}West China Hospital, Sichuan University, China\\
    \textsuperscript{\rm 4}A*STAR Institute of Advanced Intelligence and Computing, Healthcare\\
    wangao3@stu.scu.edu.cn
}

\begin{document}
\nocopyright
\maketitle

\begin{abstract}

Endoscopist gaze offers an unobtrusively acquired record of lesion-oriented visual exploration during routine clinical inspection, providing relevant supervision for tasks.
However, existing methods treat fixations as isolated spatial prompts directly associated with the target, overlooking the fact that natural gaze records a continuous visual search process in which fixations are temporally connected, spatially organized, and coupled with the visual content.
Interpreting each fixation independently based only on its local visual content overlooks the temporal evolution and spatial organization of the gaze trajectory, whereas modeling relations among fixations alone cannot reveal the visual semantics of the regions they attend to.
To this end, we propose the Gaze Exploration-Guided Segment Anything Model (GEG-SAM), a principled weakly supervised framework designed to explicitly exploit structured gaze relations and their corresponding visual cues to modulate SAM for polyp segmentation.
Specifically, GEG-SAM captures structured gaze relations by modeling the temporal dynamics and spatial topology of gaze exploration. It then associates these relational representations with fixation-aligned visual context to derive gaze-conditioned tokens, which are further injected into the frozen SAM encoder to modulate its pretrained image features.
In this way, GEG-SAM effectively exploits the rich information embedded in
gaze exploration to guide SAM toward lesion-oriented segmentation.
Extensive experiments on five public polyp segmentation datasets demonstrate
the effectiveness of our approach, improving the mean Dice by 9.30 percentage
points over the strongest weakly supervised baseline.

\end{abstract}

\section{Introduction}
\label{sec:introduction}

Medical image segmentation plays an important role in computer-assisted diagnosis and clinical decision making~\citep{qureshi2023medical}. However, obtaining pixel-level annotations is time-consuming and labor-intensive, limiting the scalability of fully supervised segmentation methods in clinical practice~\citep{shen2023survey,jiao2024learning}. Eye tracking provides a promising alternative, naturally recording visual exploration of clinicians during routine examination without requiring additional annotation procedures~\citep{bhattacharya2022radiotransformer}. Unlike explicit lesion annotations, clinical gaze arises from a continuous visual search process. Treating fixations independently based solely on local visual evidence neglects the trajectory-level context of gaze, including the temporal progression of visual search, the transitions between successive observations, and the spatial relationships among fixations~\citep{wu2025gnan,wu2026gitnet}. Therefore, jointly modeling structured gaze relations and fixation-aligned visual information is crucial for effectively exploiting natural gaze in medical image segmentation.

\begin{figure}[t]
    \centering
    \includegraphics[width=\columnwidth]{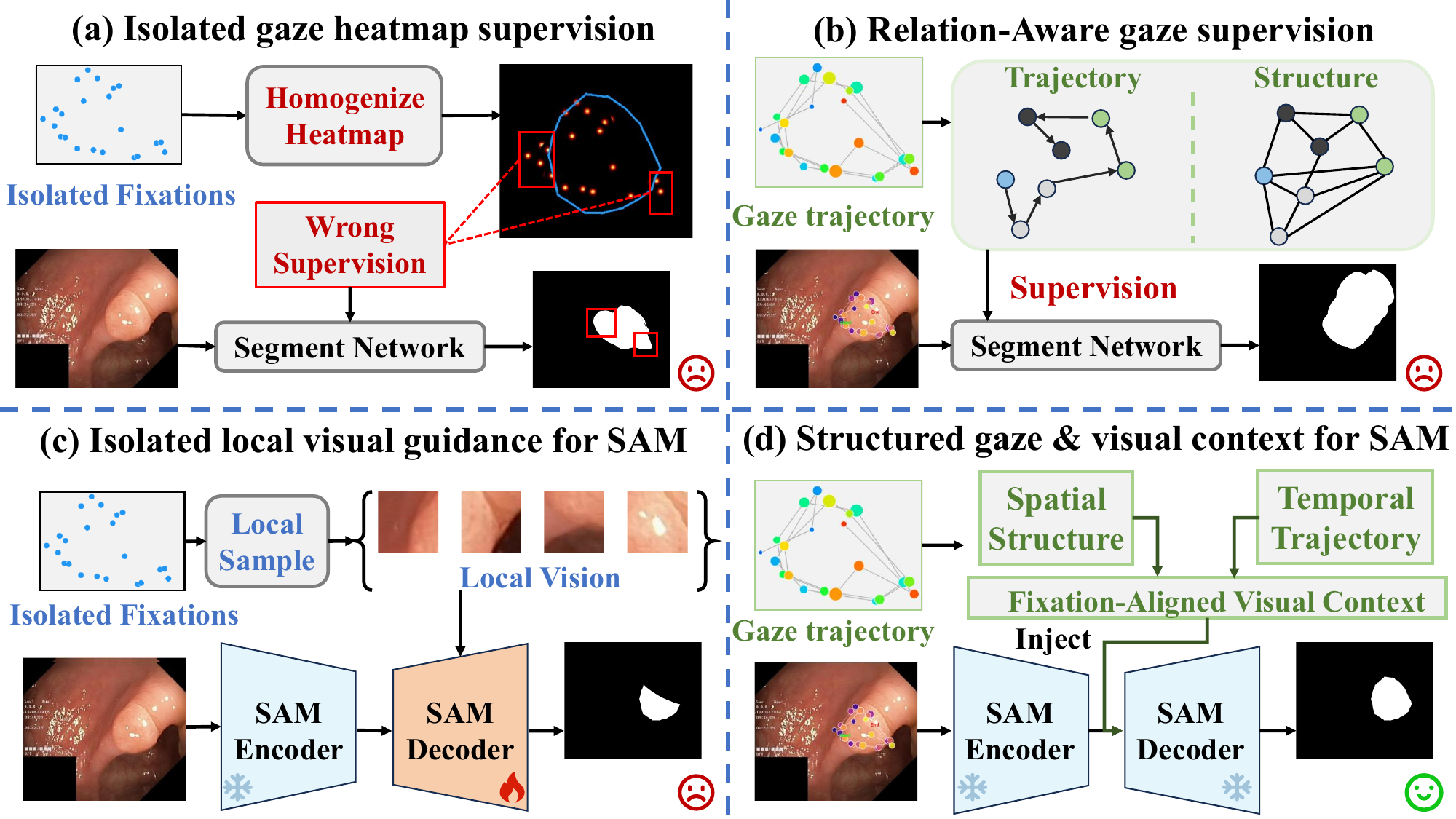}
    \caption{
    Comparison of gaze-based segmentation paradigms:
    (a) isolated gaze heatmap supervision,
    (b) relation-aware gaze supervision,
    (c) isolated local visual guidance for SAM, and
    (d) structured gaze and visual context modeling for SAM (ours).
    }
    \label{fig:Figure1}
\end{figure}

Existing gaze-supervised segmentation approaches generally fall into two categories, as illustrated in Figure~\ref{fig:Figure1}a and Figure~\ref{fig:Figure1}b. One line of work converts fixation points into gaze heatmaps and directly uses them to construct supervision~\citep{wang2022follow,zhong2024gazemedseg}. By treating fixations as isolated observations, such methods effectively homogenize their semantic roles, causing unreliable fixations to be interpreted as target-related cues and thereby compromising the accuracy and completeness of segmentation masks. The other line of work models relations among fixations by constructing graph structures to refine supervision in uncertain regions and using gaze trajectories to constrain relational representations~\citep{wu2025gnan,wu2026gitnet}. Although these methods capture temporal or structural dependencies among gaze observations, gaze is still primarily used as supervisory information or relational constraints, without explicitly associating each fixation with the specific visual content observed at its location. Consequently, the rich information embedded in gaze observations is not fully exploited. Moreover, both paradigms largely rely on feature extractors, whose representation capability is often limited by the absence of large-scale pretraining, leading to suboptimal performance.

Segmentation foundation models~\citep{kirillov2023segment,carion2025sam} pretrained on large-scale datasets exhibit strong feature extraction capabilities. Building on them, recent studies have incorporated gaze into SAM-family models for image segmentation~\citep{wang2025improving,zhao2025taming,chen2025gaze,ge2026gam}. As shown in Figure~\ref{fig:Figure1}c, these approaches explicitly couple gaze cues with pretrained image features, allowing visual information around attended locations to guide feature adaptation. However, individual fixations are still treated as isolated observations, without explicitly modeling how different gaze are temporally and spatially related. Merely associating individual fixations with their local visual features is therefore insufficient to fully interpret the information conveyed by gaze.

As illustrated in Figure~\ref{fig:Figure1}d, the relevance of each fixation should instead be understood jointly from its temporal context within the evolving exploration, its spatial relations with other observations, and the visual content encountered at the fixated location. Temporal progression and spatial organization characterize how individual observations participate in the overall search process, while the corresponding image content provides evidence of their task relevance. Jointly modeling these structured gaze relations and fixation-aligned visual information can therefore provide more informative guidance for segmentation. In this paper, we propose the Gaze Exploration-Guided Segment Anything Model
(GEG-SAM) for weakly supervised polyp segmentation. GEG-SAM first organizes
gaze exploration into structured relations by jointly characterizing its
temporal dynamics and spatial topology through a Gaze Dynamics-Structure
Encoder (GDSE). To complement these relational cues with the content actually
observed during inspection, a Gaze-Guided Semantic Token Composer (GSTC)
further couples them with fixation-aligned visual context, producing
gaze-conditioned tokens that encode both exploration structure and local image
evidence. A Gaze Feature Injection (GFI) module then transfers these tokens
into intermediate layers of the frozen SAM3 encoder, where they modulate the
pretrained image representations for lesion-oriented segmentation. Overall,
GEG-SAM enables SAM to exploit gaze as a structured visual exploration process
rather than merely as a set of isolated spatial prompts.

The main contributions include:

\begin{itemize}
    \item We propose GEG-SAM, which exploits natural gaze as structured visual exploration rather than isolated spatial observations for weakly supervised polyp segmentation.
    
    \item We introduce GDSE and GSTC to jointly model gaze dynamics, spatial organization, and fixation-aligned visual information, and inject the resulting gaze-conditioned representations into SAM3 for feature enhancement.
    
    \item Extensive experiments on five public polyp segmentation datasets demonstrate the effectiveness and strong cross-dataset generalization of GEG-SAM.
\end{itemize}

\section{Related Work}

\paragraph{Weakly Supervised Medical Image Segmentation.}
Weakly supervised medical image segmentation alleviates the reliance on dense
pixel-level annotations by learning from cheaper supervision, including
image-level labels and sparse spatial annotations such as points, bounding
boxes, and scribbles~\citep{shen2023survey,chu2025point,chi2025coarse,zhang2025helpnet}.
Recent methods further exploit segmentation foundation models to propagate
sparse prompts or refine incomplete supervision, improving segmentation under
limited annotations~\citep{zhang2024wesam,zhao2025weakpolypsam,wang2025weakmedsam}.
Related label-efficient studies have also investigated reducing annotation
uncertainty and improving pseudo-label reliability under limited supervision
or domain shifts~\citep{wang2026eauwseg,hu2026geometric}.
Although these approaches substantially reduce annotation costs, their
supervisory signals are typically specified through additional clinician
interaction and encode static spatial information. In contrast, gaze can be
naturally recorded during routine examination and provides supervision
directly from the clinician's visual exploration process without requiring a
separate annotation procedure~\citep{wang2022follow}.

\paragraph{Gaze-Based Medical Image Segmentation.}
Gaze has been incorporated into medical image segmentation in several forms.
Early approaches use gaze as auxiliary guidance to emphasize attended image
regions or enhance feature representations~\citep{wang2023eye,xie2024integrating}.
GazeMedSeg converts fixation distributions into multi-level heatmap
supervision~\citep{zhong2024gazemedseg}, while GAM uses gaze heatmaps to guide
feature- and output-level adaptation of SAM~\citep{ge2026gam}; GazeSAM instead
directly employs gaze as interactive prompts for SAM~\citep{wang2024gazesam}.
Beyond treating gaze as static spatial cues, GNAN introduces graph-based
neighbor reasoning to alleviate ambiguity in gaze supervision
~\citep{wu2025gnan}. GradTrack further emphasizes the temporal information of
gaze trajectories by incorporating fixation positions, durations, and
observation order~\citep{wang2025enjoying}, whereas GiTNet uses dynamic gaze
trajectories to constrain relational structures among image regions
~\citep{wu2026gitnet}. Despite these advances, existing methods mainly exploit
isolated fixation cues, trajectory dynamics, or structural relations from
different perspectives. The temporal evolution and spatial organization of
the complete exploration process are rarely interpreted together with the
visual evidence observed at individual fixations, leaving the information
carried by natural gaze only partially exploited.
\section{Method}
\label{sec:method}

\begin{figure*}[t]
    \centering
    \includegraphics[width=\textwidth]{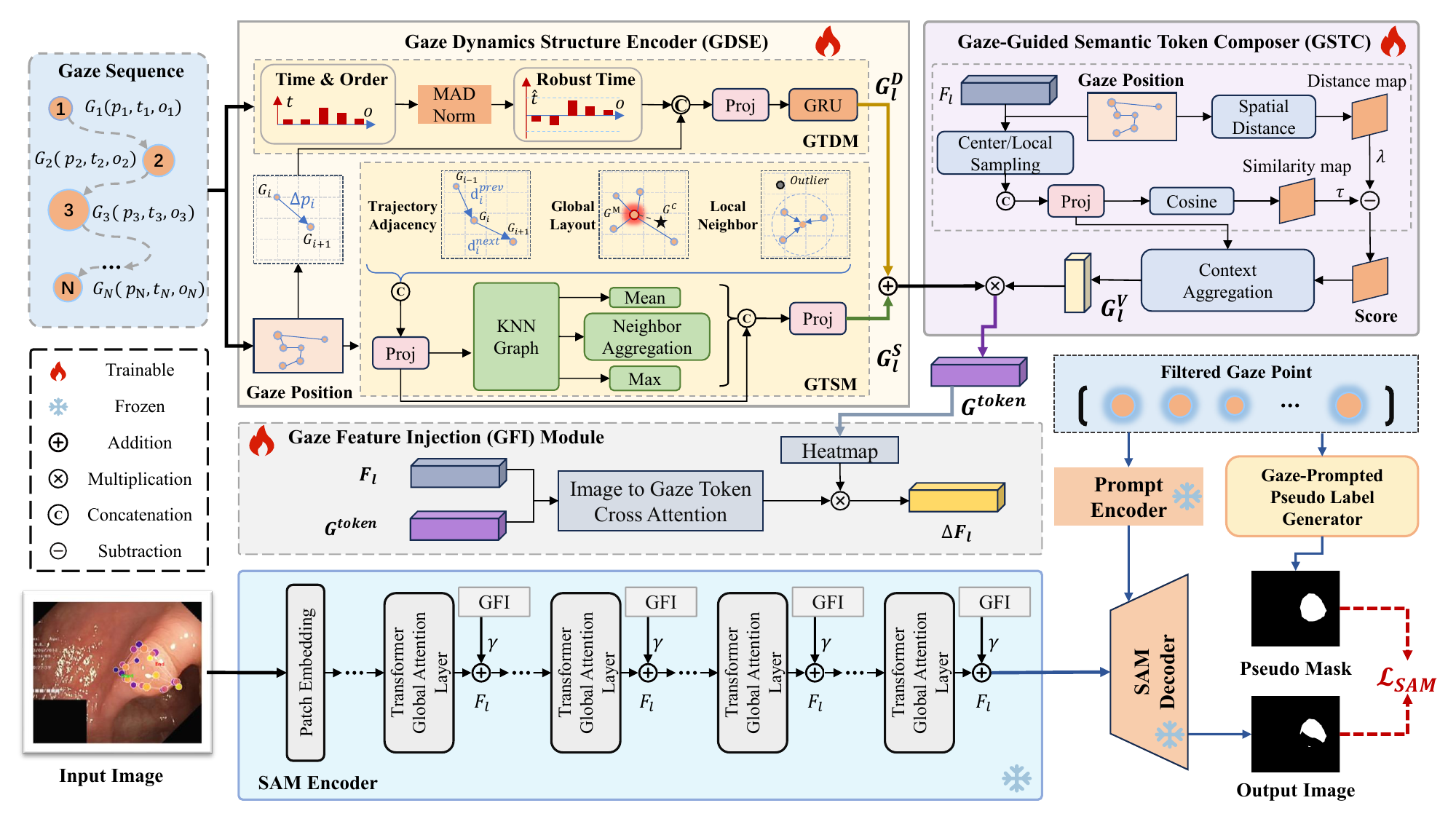}
    \caption{
    Overview of \textbf{GEG-SAM}.
    GDSE characterizes the dynamics and spatial organization of the gaze
    trajectory, while GSTC associates these relational cues with
    fixation-aligned image evidence to compose semantic gaze tokens.
    GFI further introduces the resulting gaze guidance into selected
    layers of the frozen SAM3 encoder.
    Duration-filtered fixations are also used as positive prompts and
    to derive pseudo masks for weak supervision.
    }
    \label{fig:Figure2}
\end{figure*}

\subsection{Overview}
\label{sec:overview}

Given an endoscopic image and its synchronized gaze sequence $\mathcal{G}=\{(p_i,t_i,o_i)\}_{i=1}^{N}$, where $p_i$, $t_i$, and $o_i$ denote the fixation position,
duration, and observation order of the $i$-th fixation, respectively. We construct GEG-SAM as illustrated in Figure~\ref{fig:Figure2}. The input image is first processed by the SAM3 image encoder to obtain hierarchical image representations. Meanwhile, the complete gaze sequence is fed into gaze dynamics-structure encoder (GDSE), which models fixation-level relations from the perspectives of temporal dynamics and spatial topology. At selected encoder layers, gaze-guided semantic token composer (GSTC) extracts fixation-aligned visual information from the current image features and combines it with the GDSE representations to compose semantic gaze representations. Gaze feature injection (GFI) module then maps these gaze representations back to the dense image space and applies them to the corresponding encoder features through residual updates, after which the enhanced representations are propagated through the subsequent Transformer layers.

The gaze-enhanced image representations are finally passed to the SAM3 decoder together with positive point prompts derived from duration-filtered fixations to produce the segmentation mask. In addition, the same filtered fixations are used to prompt SAM3 for pseudo-mask generation, providing pixel-level supervision for model training.

\subsection{Gaze Dynamics-Structure Encoder}
\label{sec:gdse}

In natural gaze exploration, the role of an individual fixation is difficult
to determine from its spatial location alone. However, coherent inspection
and incidental observations exhibit different temporal dynamics and spatial
organization within the gaze trajectory. Across the exploration process,
these relational differences provide discriminative cues for separating
coherent observations from isolated ones. We design a Gaze
Dynamics-Structure Encoder (GDSE) to extract such relational cues. As
illustrated in Figure~\ref{fig:Figure2}, GDSE consists of Gaze Temporal
Dynamics Modeling (GTDM) to model the sequential evolution of gaze behavior
and Gaze Topology Structure Modeling (GTSM) to characterize the spatial
organization among fixations.

\paragraph{Gaze Temporal Dynamics Modeling.}
Fixation duration reflects how persistently a region is observed, while
observation order and inter-fixation transitions characterize how visual
attention evolves throughout the exploration process. Among these cues,
raw fixation duration is particularly sensitive to scale variation across
gaze sequences. We therefore first apply logarithmic transformation followed
by median absolute deviation (MAD) normalization:

\begin{equation}
\widehat{t}_i
=
\frac{
\log(1+t_i)
-
\operatorname{median}_{j}
\left(
\log(1+t_j)
\right)
}{
\operatorname{MAD}
\left(
\{\log(1+t_j)\}_{j=1}^{N}
\right)
}.
\label{eq:mad_norm}
\end{equation}

This normalization preserves relative duration variation within each
exploration while reducing sensitivity to absolute scale and unusually long
fixations. Duration alone, however, only characterizes the persistence of an
individual observation. To describe how that observation participates in the
ongoing exploration, we further incorporate its position in the sequence and
the change from the preceding fixation:

\begin{equation}
\mathbf{q}_{i}^{D}
=
\operatorname{Concat}
\left[
\widehat{t}_i,\,
\bar{o}_i,\,
\Delta\widehat{t}_i,\,
\Delta\mathbf{p}_i
\right],
\label{eq:temporal_descriptor}
\end{equation}
where $\bar{o}_i$ denotes the normalized observation order,
$\Delta\widehat{t}_i$ represents the duration change, and
$\Delta\mathbf{p}_i$ characterizes the spatial transition from the preceding
fixation. While $\mathbf{q}_{i}^{D}$ captures the local behavioral change
associated with the current observation, visual exploration evolves through
a sequence of such transitions whose effects accumulate over time. We
therefore employ a unidirectional GRU to progressively integrate preceding
gaze dynamics while preserving the forward order of inspection, which is defined as:

\begin{equation}
\mathbf{G}_{i}^{D}
=
\operatorname{GRU}
\left(
\operatorname{Proj}
\left(
\mathbf{q}_{i}^{D}
\right),
\mathbf{G}_{i-1}^{D}
\right),
\label{eq:gtdm}
\end{equation}
where $\operatorname{Proj}(\cdot)$ denotes projection function, $\mathbf{G}_{i}^{D}$ is temporal dynamics representation of the
$i$-th fixation. The recurrent state accumulates the preceding behavioral
evolution, allowing each fixation to be represented in the context of how
visual attention has progressed up to the current observation.

\paragraph{Gaze Topology Structure Modeling.}
Temporal dynamics characterize how an observation emerges along the scanpath,
but they do not reveal how it is situated within the spatial organization of
the complete gaze trajectory. A fixation may lie within a coherent inspection
region, occur along a transition between regions, or deviate from the dominant
exploration pattern. We therefore first characterize its fixation-level
geometry through trajectory adjacency, global layout, and local neighborhood
relations:
\begin{equation}
\begin{aligned}
\mathbf{r}_{i}^{A}
&=
\left[
d_{i}^{\mathrm{prev}},
d_{i}^{\mathrm{next}}
\right],\\
\mathbf{r}_{i}^{G}
&=
\left[
\mathbf{p}_{i}-\mathbf{p}^{C},
d_{i}^{C},
d_{i}^{M}
\right],\\
\mathbf{r}_{i}^{L}
&=
\left[
d_{i}^{\mathrm{near}},
\bar{d}_{i}^{K},
\rho_i
\right],\\
\mathbf{h}_{i}^{S}
&=
\operatorname{Proj}
\left(
\left[
\mathbf{r}_{i}^{A},
\mathbf{r}_{i}^{G},
\mathbf{r}_{i}^{L}
\right]
\right).
\end{aligned}
\label{eq:gaze_spatial_relations}
\end{equation}
where $d_{i}^{\mathrm{prev}}$ and $d_{i}^{\mathrm{next}}$ denote the distances to the preceding and succeeding fixations. $\mathbf{p}^{C}$ is the gaze centroid, while $d_{i}^{C}$ and $d_{i}^{M}$ denote the distances to the centroid and medoid. $d_{i}^{\mathrm{near}}$, $\bar{d}_{i}^{K}$, and
$\rho_i$ respectively denote the nearest-neighbor distance, mean $K$-NN
distance, and local fixation density. These cues are projected into the
fixation-level structural representation $\mathbf{h}_{i}^{S}$.

Fixation-level geometry alone does not capture whether nearby observations
form a consistent spatial organization. We therefore construct a
$K$-nearest-neighbor (KNN) graph according to fixation positions and
aggregate structural information from spatially neighboring observations:

\begin{equation}
\mathbf{n}_{i}^{S}
=
\frac{1}{|\mathcal{N}_{i}^{K}|}
\sum_{j\in\mathcal{N}_{i}^{K}}
\mathbf{h}_{j}^{S},
\label{eq:neighbor_aggregation}
\end{equation}
where $\mathcal{N}_{i}^{K}$ denotes the $K$ spatially nearest fixations.
Unlike trajectory adjacency, which follows observation order, this
neighborhood relation connects observations that may be temporally separated
but spatially coherent.

Finally, local organization must be interpreted with respect to the complete
exploration pattern. We summarize the gaze set through mean and max pooling,
obtaining $\mathbf{g}_{\mathrm{mean}}^{S}$ and
$\mathbf{g}_{\mathrm{max}}^{S}$, and integrate fixation-level, neighborhood,
and global structural information as:

\begin{equation}
\mathbf{G}_{i}^{S}
=
\operatorname{Proj}
\left(
\left[
\mathbf{h}_{i}^{S},
\mathbf{n}_{i}^{S},
\mathbf{g}_{\mathrm{mean}}^{S},
\mathbf{g}_{\mathrm{max}}^{S}
\right]
\right).
\label{eq:gtsm}
\end{equation}

The resulting $\mathbf{G}_{i}^{S}$ progressively places each fixation within
its own geometry, its spatial neighborhood, and the organization of the
complete gaze trajectory.

\subsection{Gaze-guided Semantic Token Composer}
\label{sec:gstc}

GDSE characterizes how each fixation participates in visual exploration, but
relational context alone cannot determine what is actually observed at that
location. We therefore introduce the Gaze-Guided Semantic Token Composer
(GSTC) to associate gaze context with fixation-aligned image evidence and
resolve the visual semantics of individual observations.

Given the feature map $\mathbf{F}_l$ at the $l$-th selected encoder layer, we
sample both the feature at each fixation and its locally averaged context.
Their discrepancy captures how the observed location differs from its
surroundings:

\begin{equation}
\begin{aligned}
\mathbf{f}_{l,i}^{c}
&=
\operatorname{Sample}
(\mathbf{F}_l,\mathbf{p}_i),\\
\mathbf{f}_{l,i}^{r}
&=
\operatorname{Sample}
(\operatorname{AvgPool}(\mathbf{F}_l),\mathbf{p}_i),\\
\mathbf{v}_{l,i}
&=
\operatorname{Proj}_{V}
\left(
\left[
\mathbf{f}_{l,i}^{c};
\mathbf{f}_{l,i}^{r};
\left|
\mathbf{f}_{l,i}^{c}-\mathbf{f}_{l,i}^{r}
\right|
\right]
\right).
\end{aligned}
\label{eq:visual_sampling}
\end{equation}

The visual evidence from a single fixation can remain ambiguous, particularly
when lesion and surrounding tissue exhibit similar local appearance. We
therefore further seek supporting evidence from other fixations across the
complete visual exploration rather than interpreting each observation
independently. Specifically, for fixations $i$ and $j$, we measure their
relation by combining visual similarity with spatial proximity:

\begin{equation}
s_{l,ij}
=
\frac{
\operatorname{cos}
(\mathbf{v}_{l,i},\mathbf{v}_{l,j})
}{\tau}
-
\lambda
\|\mathbf{p}_{i}-\mathbf{p}_{j}\|_{2}^{2},
\qquad i\neq j ,
\label{eq:relation_score}
\end{equation}
where $\tau$ controls the similarity scale and $\lambda$ regulates the spatial
constraint. The top-$K$ related observations are then aggregated according to
their normalized relation scores:

\begin{equation}
\mathbf{c}_{l,i}
=
\sum_{j\in\mathcal{R}_{l,i}^{K}}
\frac{\exp(s_{l,ij})}
{\sum_{k\in\mathcal{R}_{l,i}^{K}}\exp(s_{l,ik})}
\mathbf{v}_{l,j},
\label{eq:context_aggregation}
\end{equation}
where $\mathcal{R}_{l,i}^{K}$ denotes the selected supporting observations.
The fixation-aligned evidence, aggregated context, and their discrepancy are
fused as
\begin{equation}
\mathbf{G}_{l,i}^{V}
=
\operatorname{Proj}
\left(
\left[
\mathbf{v}_{l,i};
\mathbf{c}_{l,i};
\left|
\mathbf{v}_{l,i}-\mathbf{c}_{l,i}
\right|
\right]
\right),
\label{eq:visual_fusion}
\end{equation}
allowing each observation to incorporate visual evidence shared by related fixations. Finally, we combine the visual evidence with the temporal dynamics and
spatial structure encoded by GDSE to describe how the current observation is
situated within the complete visual exploration process:

\begin{equation}
\mathbf{G}_{l,i}^{\mathrm{token}}
=
\operatorname{LN}
\left(
\mathbf{G}_{l,i}^{V}
\odot
\left[
\operatorname{Proj}_{D}
\left(
\mathbf{G}_{i}^{D}
\right)
+
\operatorname{Proj}_{S}
\left(
\mathbf{G}_{i}^{S}
\right)
\right]
\right).
\label{eq:gaze_token}
\end{equation}

The resulting $\mathbf{G}_{l,i}^{\mathrm{token}}$ jointly encodes what is
observed at the fixation and how that observation participates in the
ongoing gaze exploration, providing semantic gaze guidance for subsequent
feature injection.

\subsection{Gaze Feature Injection}
\label{sec:gfi}

GSTC produces fixation-level semantic representations, which must be
introduced into the dense image features without disrupting the pretrained
SAM3 representation. We therefore insert GFI after selected global-attention
blocks of the frozen image encoder. To adapt the spatial influence of each
fixation to its semantic representation, GFI predicts a fixation-specific
spatial scale and uses it to construct a gaze-conditioned heatmap:
\begin{equation}
\begin{aligned}
\sigma_{l,i}
&=
\operatorname{Proj}_{\sigma}
\left(
\mathbf{G}_{l,i}^{\mathrm{token}}
\right),\\
\mathbf{H}_{l}(\mathbf{u})
&=
\max_{i}
\exp
\left(
-\frac{
\|\mathbf{u}-\mathbf{p}_{i}\|_2^{2}
}{
2\sigma_{l,i}^{2}
}
\right).
\end{aligned}
\label{eq:adaptive_heatmap}
\end{equation}
where $\mathbf{u}$ denotes a spatial location on the feature map, and
$\sigma_{l,i}$ determines the spatial influence of the $i$-th fixation, allowing different semantic observations to induce different
spatial extents of gaze guidance. The resulting heatmap constrains this
guidance to fixation-related regions, while cross-attention transfers the
semantic gaze representations to the dense image features:

\begin{equation}
\widetilde{\mathbf{F}}_{l}
=
\mathbf{F}_{l}
+
\gamma_l
\mathbf{H}_{l}
\odot
\operatorname{CrossAttn}
\left(
\mathbf{F}_{l},
\mathbf{G}_{l}^{\mathrm{token}}
\right),
\label{eq:gfi}
\end{equation}
where $\gamma_l$ is a learnable residual scale. The updated representation
$\widetilde{\mathbf{F}}_{l}$ is subsequently propagated through the remaining
frozen encoder layers.

\subsection{Loss Function}
\label{sec:loss}

We derive mask-level supervision from the same gaze observations without using manual masks during training. Fixations with sufficiently long duration are retained as positive prompts and passed to the frozen SAM3 model to obtain gaze-prompted pseudo masks. Gaussian smoothing is applied to suppress isolated artifacts, yielding the supervision mask $\widetilde{\mathbf{M}}$. 

The filtered fixations are also used as positive prompts to the frozen prompt encoder during GEG-SAM training, whereas the complete fixation sequence is preserved for gaze feature modeling. Only GDSE, GSTC, and GFI are optimized. Following the standard segmentation objective, we combine binary cross-entropy loss and soft Dice loss~\citep{kullback1951information,milletari2016v}. The
overall objective is defined as:

\begin{equation}
\mathcal{L}_{\mathrm{SAM}}
=
\mathcal{L}_{\mathrm{BCE}}
\left(
\widehat{\mathbf{M}},
\widetilde{\mathbf{M}}
\right)
+
\mathcal{L}_{\mathrm{Dice}}
\left(
\widehat{\mathbf{M}},
\widetilde{\mathbf{M}}
\right),
\label{eq:loss}
\end{equation}
where $\widehat{\mathbf{M}}$ denotes the segmentation predicted from the
gaze-enhanced image representation and the filtered positive prompts.
\section{Experiments}
\label{sec:experiments}

\subsection{Experimental Setup}

\paragraph{Dataset.}
Experiments are conducted on five public colonoscopy datasets, including Kvasir-SEG~\citep{jha2019kvasir}, CVC-ClinicDB~\citep{bernal2015wm}, CVC-ColonDB~\citep{tajbakhsh2015automated}, CVC-300~\citep{vazquez2017benchmark}, and ETIS~\citep{silva2014toward}. Following PraNet~\citep{fan2020pranet}, 1,450 images from Kvasir-SEG and CVC-ClinicDB are used for training, while the remaining 162 images are used for in-domain testing. CVC-ColonDB, CVC-300, and ETIS, containing 636 images in total, are used as unseen test sets. Gaze annotations for Kvasir-SEG are adopted from GazeMedSeg~\citep{zhong2024gazemedseg}. For the remaining datasets, following the annotation protocol of GazeMedSeg, we collected gaze data from experienced endoscopists using a Tobii Pro Fusion eye tracker at a sampling rate of 250 Hz.

\paragraph{Implementation Details.}
Segmentation performance is evaluated using Dice and ASSD. All images are resized to $1008\times1008$. Given its strong feature representation capability, we adopt SAM3 as the primary backbone of GEG-SAM. For a fair comparison, we additionally evaluate GEG-SAM with the SAM backbones under the same experimental setting. The SAM3 backbones are kept frozen, and only the proposed modules inserted into their image encoders are optimized. The model is trained for 20 epochs using Adam with an initial learning rate of $1\times10^{-2}$, and a batch size of 1. For the main comparison in Table 1, our results are averaged over three independent runs with different fixed random seeds, while all other experiments are conducted using a single fixed seed. All methods are implemented in PyTorch and trained on a single NVIDIA GeForce RTX 4090 GPU.

\subsection{Comparison with State-of-the-Art Methods}
\label{sec:sota_comparison}

\paragraph{Quantitative Results.}
\MainResultsTable

Table~\ref{tab:main_results} reports the quantitative comparison results.
Compared with gaze-supervised and weakly supervised methods, GEG-SAM with SAM3 achieves a mean Dice of $86.43\%$ and an ASSD of $18.45$ pixels. It improves over GAM by $10.14$ percentage points in mean Dice while reducing ASSD by $2.37$ pixels, and surpasses WeSAM by $9.30$ percentage points with a further $0.3$-pixel reduction in ASSD. These gains demonstrate the benefit of jointly exploiting structured gaze relations and fixation-aligned visual evidence.

Compared with fully supervised and foundation-model baselines, GEG-SAM achieves an average Dice of $85.26\%$ across the three cross-domain datasets, outperforming Polyp-PVT by $1.67$ percentage points, while also exceeding SAMed-2 by $3.78$ percentage points in mean Dice. Medical foundation models benefit from medical-domain pretraining or task-specific adaptation that includes data from the evaluated domains, while oracle box prompting provides explicit target localization, which explains their advantage in some settings. Nevertheless, GEG-SAM remains competitive without relying on dense annotations or explicit target boxes.

We further evaluate GEG-SAM with SAM ViT-B, ViT-L, and ViT-H backbones, which achieve mean Dice scores of $85.23\%$, $85.99\%$, and $85.94\%$, respectively. The consistently strong performance across different SAM backbones indicates that the proposed gaze modeling strategy is not specific to SAM3 and generalizes well across backbone scales.

\paragraph{Qualitative Results.}
\begin{figure*}[t]
    \centering
    \includegraphics[width=\textwidth]{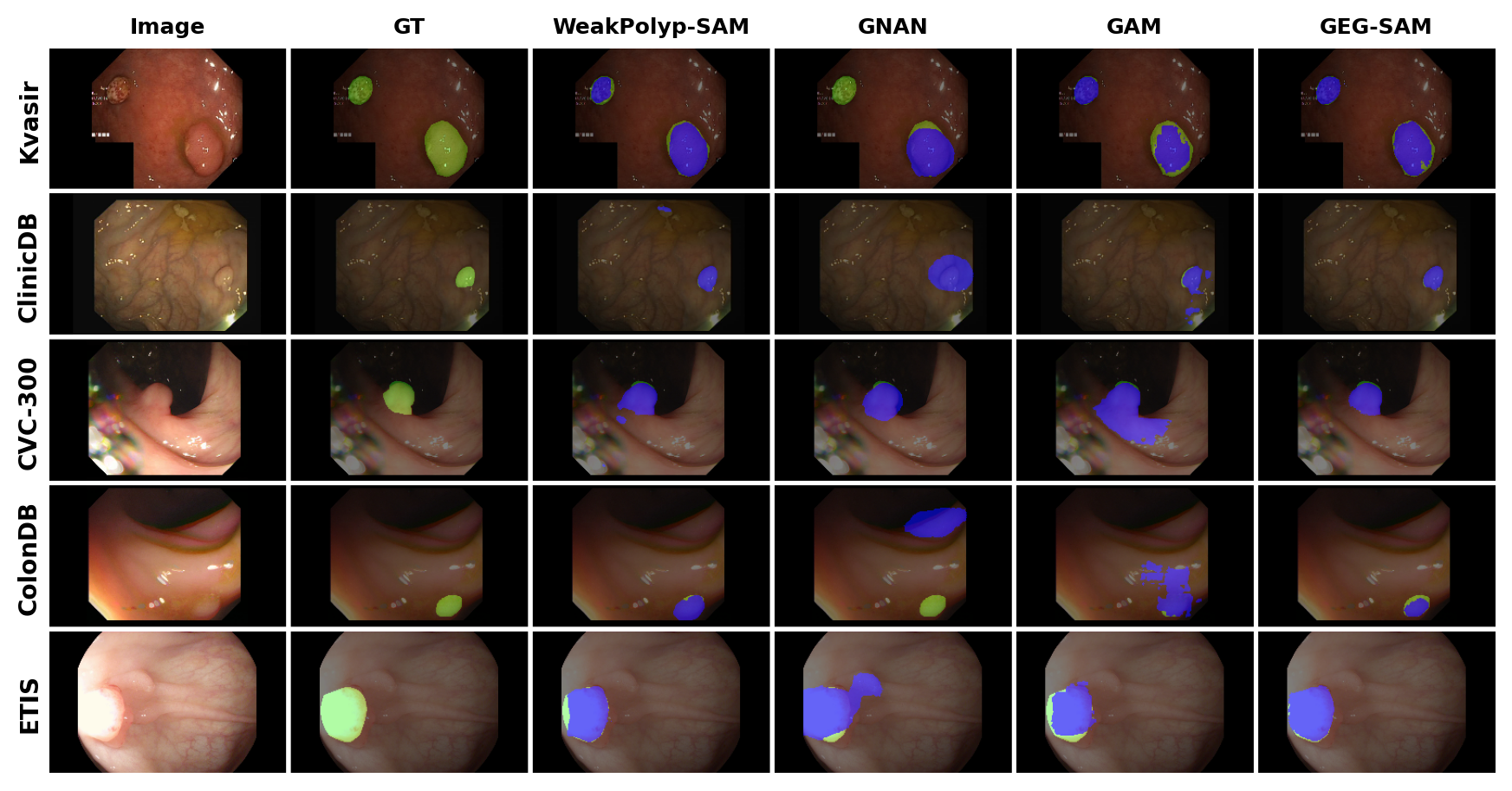}
    \caption{Qualitative comparisons with representative methods, including WeakPolyp-SAM, GNAN, and GAM, across five polyp segmentation datasets. Green and blue denote the ground-truth and predicted masks, respectively.}
    \label{fig:qualitative_results}
\end{figure*}

Representative examples are shown in Figure~\ref{fig:qualitative_results}.
Existing methods often produce incomplete or fragmented masks under ambiguous
lesion appearances and boundaries, whereas GEG-SAM yields more compact and
coherent predictions that better match the ground truth. This further shows
that structured gaze modeling helps SAM focus on lesion-relevant observations
while suppressing distracting fixations and similar background regions.

\subsection{Ablation Study}
\label{sec:ablation}

\paragraph{Component Analysis.}
\ComponentAblationTable

As shown in Table~\ref{tab:component_ablation}, introducing GDSE and GSTC
individually improves the mean Dice from $80.18\%$ to $83.99\%$ and $85.49\%$,
respectively, while combining them further increases it to $86.12\%$.
When GFI is removed, the gaze-conditioned tokens are directly added to the image features at their corresponding fixation locations.
Adding GFI achieves the best mean Dice of $86.60\%$ and cross-domain Dice of
$85.50\%$. These results verify the effectiveness of structured gaze modeling,
fixation-aligned visual context, and their integration into SAM features.

\paragraph{Effect of Gaze Dynamics-Structure Encoder.}

\begin{figure}[t]
    \centering
    \includegraphics[width=\columnwidth]{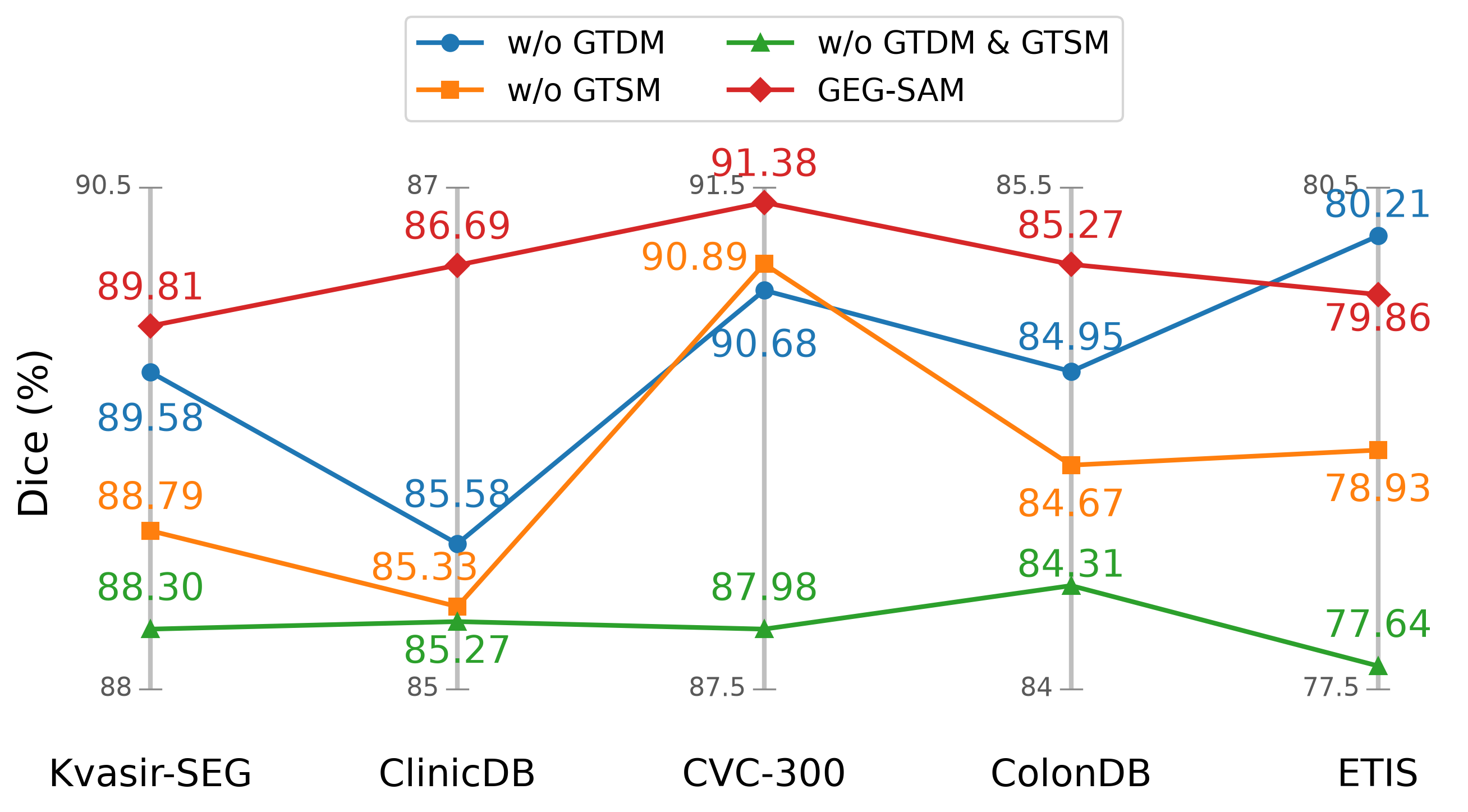}
    \caption{Ablation of temporal dynamics and spatial topology in GDSE
    across five polyp segmentation datasets.}
    \label{fig:trajectory_cue_ablation}
\end{figure}

Figure~\ref{fig:trajectory_cue_ablation} further evaluates the two branches of
GDSE. Using only spatial topology or temporal dynamics yields mean Dice scores
of $86.20\%$ and $85.72\%$, respectively, while removing both decreases it to
$84.70\%$. Joint modeling achieves the best result of $86.60\%$, confirming
that both temporal and spatial relations contribute to gaze representation.

\paragraph{Effect of Gaze-Guided Semantic Token Composer.}

\begin{figure}[t]
    \centering
    \includegraphics[width=\linewidth]{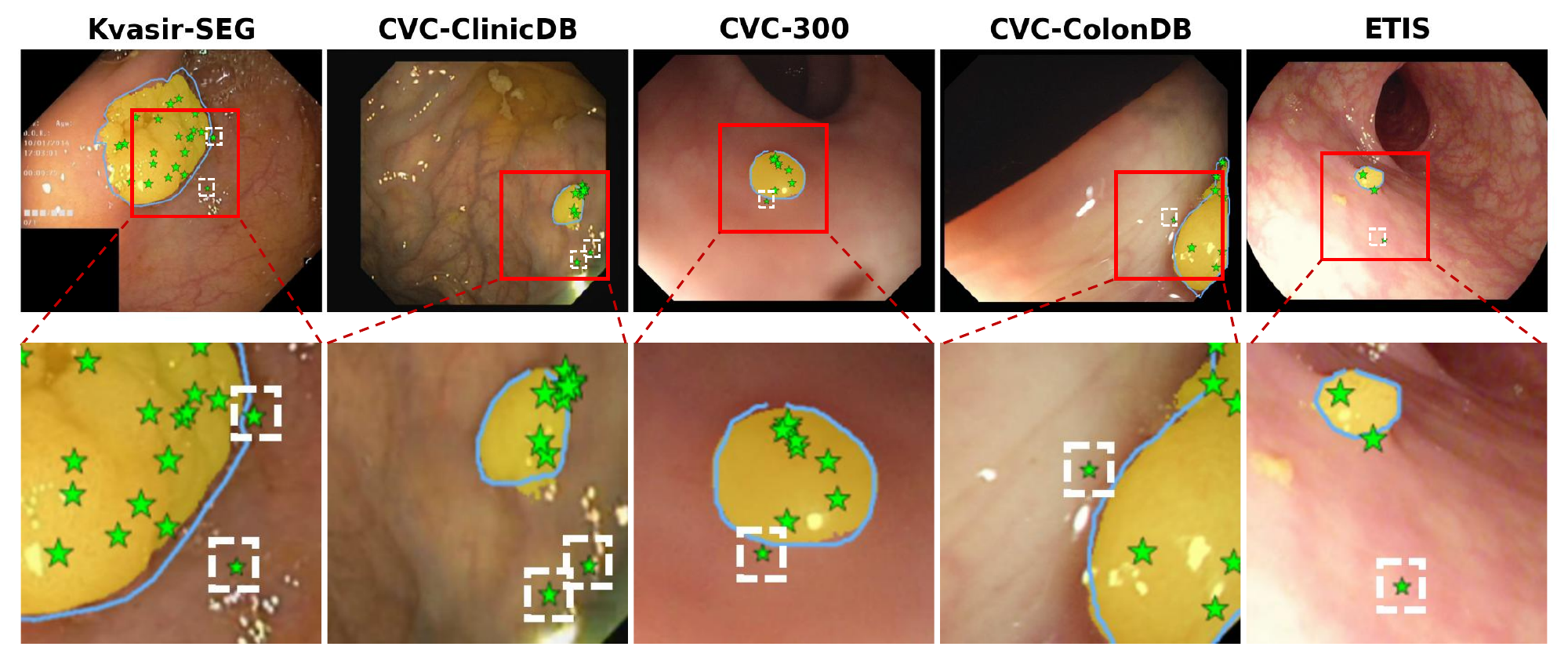}
    \caption{Visualization of predicted spatial scales $\sigma$ for gaze
    points. Blue contours, yellow regions, and green stars denote ground truth,
    predictions, and gaze points, respectively.}
    \label{fig:gstc_sigma}
\end{figure}

As shown in Fig.~\ref{fig:gstc_sigma}, lesion-related gaze points generally
receive larger spatial scales, whereas gaze points outside the lesion are
assigned smaller ones. This indicates that the gaze-conditioned representations
emphasize relevant observations while limiting the influence of background
fixations.

\section{Conclusion}
We presented GEG-SAM for gaze-guided weakly supervised polyp segmentation. Our results show that gaze is more effective when interpreted as a structured visual exploration process rather than a collection of isolated point prompts. By exploiting the structure relations among fixations and associated visual context, GEG-SAM provides reliable guidance for adapting a frozen SAM3 and achieves strong performance across five polyp datasets. Future work will extend GEG-SAM to endoscopic video segmentation by jointly modeling continuous gaze exploration and inter-frame visual dynamics, enabling gaze guidance to evolve coherently across time.

\clearpage

\bibliography{ref}

@article{qureshi2023medical,
  title={Medical image segmentation using deep semantic-based methods: A review of techniques, applications and emerging trends},
  author={Qureshi, Imran and Yan, Junhua and Abbas, Qaisar and Shaheed, Kashif and Riaz, Awais Bin and Wahid, Abdul and Khan, Muhammad Waseem Jan and Szczuko, Piotr},
  journal={Information Fusion},
  volume={90},
  pages={316--352},
  year={2023},
  publisher={Elsevier}
}

@article{shen2023survey,
  title={A survey on label-efficient deep image segmentation: Bridging the gap between weak supervision and dense prediction},
  author={Shen, Wei and Peng, Zelin and Wang, Xuehui and Wang, Huayu and Cen, Jiazhong and Jiang, Dongsheng and Xie, Lingxi and Yang, Xiaokang and Tian, Qi},
  journal={IEEE transactions on pattern analysis and machine intelligence},
  volume={45},
  number={8},
  pages={9284--9305},
  year={2023},
  publisher={IEEE}
}

@article{jiao2024learning,
  title={Learning with limited annotations: a survey on deep semi-supervised learning for medical image segmentation},
  author={Jiao, Rushi and Zhang, Yichi and Ding, Le and Xue, Bingsen and Zhang, Jicong and Cai, Rong and Jin, Cheng},
  journal={Computers in Biology and Medicine},
  volume={169},
  pages={107840},
  year={2024},
  publisher={Elsevier}
}

@inproceedings{bhattacharya2022radiotransformer,
  title={RadioTransformer: A cascaded global-focal transformer for visual attention--guided disease classification},
  author={Bhattacharya, Moinak and Jain, Shubham and Prasanna, Prateek},
  booktitle={European Conference on Computer Vision},
  pages={679--698},
  year={2022},
  organization={Springer}
}

@article{wang2022follow,
  title={Follow my eye: Using gaze to supervise computer-aided diagnosis},
  author={Wang, Sheng and Ouyang, Xi and Liu, Tianming and Wang, Qian and Shen, Dinggang},
  journal={IEEE Transactions on Medical Imaging},
  volume={41},
  number={7},
  pages={1688--1698},
  year={2022},
  publisher={IEEE}
}

@inproceedings{kirillov2023segment,
  title={Segment anything},
  author={Kirillov, Alexander and Mintun, Eric and Ravi, Nikhila and Mao, Hanzi and Rolland, Chloe and Gustafson, Laura and Xiao, Tete and Whitehead, Spencer and Berg, Alexander C and Lo, Wan-Yen and others},
  booktitle={Proceedings of the IEEE/CVF international conference on computer vision},
  pages={4015--4026},
  year={2023}
}

@article{carion2025sam,
  title={Sam 3: Segment anything with concepts},
  author={Carion, Nicolas and Gustafson, Laura and Hu, Yuan-Ting and Debnath, Shoubhik and Hu, Ronghang and Suris, Didac and Ryali, Chaitanya and Alwala, Kalyan Vasudev and Khedr, Haitham and Huang, Andrew and others},
  journal={arXiv preprint arXiv:2511.16719},
  year={2025}
}

@article{wang2025improving,
  title={Improving self-supervised medical image pre-training by early alignment with human eye gaze information},
  author={Wang, Sheng and Zhao, Zihao and Shen, Zhenrong and Wang, Bin and Wang, Qian and Shen, Dinggang},
  journal={IEEE Transactions on Medical Imaging},
  volume={44},
  number={10},
  pages={4063--4072},
  year={2025},
  publisher={IEEE}
}

@inproceedings{zhao2025taming,
  title={Taming Masked Image Modeling for Chest X-ray Diagnosis by Incorporating Clinical Visual Priors},
  author={Zhao, Zihao and Wang, Mei and Cui, Zhiming and Wang, Sheng and Zhou, Qian and Fan, Li and Wang, Qian and Shen, Dinggang},
  booktitle={International Conference on Information Processing in Medical Imaging},
  pages={157--171},
  year={2025},
  organization={Springer}
}

@article{chen2025gaze,
  title={From gaze to insight: Bridging human visual attention and vision language model explanation for weakly-supervised medical image segmentation},
  author={Chen, Jingkun and Duan, Haoran and Zhang, Xiao and Gao, Boyan and Grau, Vicente and Han, Jungong},
  journal={IEEE Transactions on Medical Imaging},
  year={2025},
  publisher={IEEE}
}

@article{chu2025point,
  title={Point-annotation supervision for robust 3D pulmonary infection segmentation by CT-based cascading deep learning},
  author={Chu, Yuetan and Wang, Jianpeng and Xiong, Yaxin and Gao, Yuan and Liu, Xin and Luo, Gongning and Gao, Xin and Zhao, Mingyan and Huang, Chao and Qiu, Zhaowen and others},
  journal={Computers in Biology and Medicine},
  volume={187},
  pages={109760},
  year={2025},
  publisher={Elsevier}
}

@article{chi2025coarse,
  title={Coarse for fine: Bounding box supervised thyroid ultrasound image segmentation using spatial arrangement and hierarchical prediction consistency},
  author={Chi, Jianning and Lin, Geng and Li, Zelan and Zhang, Wenjun and Chen, Jia-hui and Huang, Ying},
  journal={IEEE Journal of Biomedical and Health Informatics},
  volume={29},
  number={6},
  pages={4186--4199},
  year={2025},
  publisher={IEEE}
}

@article{zhang2025helpnet,
  title={HELPNet: Hierarchical perturbations consistency and entropy-guided ensemble for scribble supervised medical image segmentation},
  author={Zhang, Xiao and Wu, Shaoxuan and Zhang, Peilin and Jin, Zhuo and Xiong, Xiaosong and Bu, Qirong and Chen, Jingkun and Feng, Jun},
  journal={Medical Image Analysis},
  pages={103719},
  year={2025},
  publisher={Elsevier}
}

@inproceedings{wang2023eye,
  title={Eye-guided dual-path network for multi-organ segmentation of abdomen},
  author={Wang, Chong and Zhang, Daoqiang and Ge, Rongjun},
  booktitle={International Conference on Medical Image Computing and Computer-Assisted Intervention},
  pages={23--32},
  year={2023},
  organization={Springer}
}

@article{xie2024integrating,
  title={Integrating eye tracking with grouped fusion networks for semantic segmentation on mammogram images},
  author={Xie, Jiaming and Zhang, Qing and Cui, Zhiming and Ma, Chong and Zhou, Yan and Wang, Wenping and Shen, Dinggang},
  journal={IEEE Transactions on Medical Imaging},
  volume={44},
  number={2},
  pages={868--879},
  year={2024},
  publisher={IEEE}
}

@inproceedings{wang2024gazesam,
  title={Gazesam: Interactive image segmentation with eye gaze and segment anything model},
  author={Wang, Bin and Aboah, Armstrong and Zhang, Zheyuan and Pan, Hongyi and Bagci, Ulas},
  booktitle={Gaze Meets Machine Learning Workshop},
  pages={254--265},
  year={2024},
  organization={PMLR}
}

@inproceedings{wang2025enjoying,
  title={Enjoying Information Dividend: Gaze Track-Based Medical Weakly Supervised Segmentation},
  author={Wang, Zhisong and Ye, Yiwen and Chen, Ziyang and Xia, Yong},
  booktitle={International Conference on Medical Image Computing and Computer-Assisted Intervention},
  pages={202--212},
  year={2025},
  organization={Springer}
}

@article{kullback1951information,
  title={On information and sufficiency},
  author={Kullback, Solomon and Leibler, Richard A},
  journal={The annals of mathematical statistics},
  volume={22},
  number={1},
  pages={79--86},
  year={1951},
  publisher={JSTOR}
}

@inproceedings{milletari2016v,
  title={V-net: Fully convolutional neural networks for volumetric medical image segmentation},
  author={Milletari, Fausto and Navab, Nassir and Ahmadi, Seyed-Ahmad},
  booktitle={2016 fourth international conference on 3D vision (3DV)},
  pages={565--571},
  year={2016},
  organization={Ieee}
}

@inproceedings{jha2019kvasir,
  title={Kvasir-seg: A segmented polyp dataset},
  author={Jha, Debesh and Smedsrud, Pia H and Riegler, Michael A and Halvorsen, P{\aa}l and De Lange, Thomas and Johansen, Dag and Johansen, H{\aa}vard D},
  booktitle={International conference on multimedia modeling},
  pages={451--462},
  year={2019},
  organization={Springer}
}

@article{bernal2015wm,
  title={WM-DOVA maps for accurate polyp highlighting in colonoscopy: Validation vs. saliency maps from physicians},
  author={Bernal, Jorge and S{\'a}nchez, F Javier and Fern{\'a}ndez-Esparrach, Gloria and Gil, Debora and Rodr{\'\i}guez, Cristina and Vilari{\~n}o, Fernando},
  journal={Computerized medical imaging and graphics},
  volume={43},
  pages={99--111},
  year={2015},
  publisher={Elsevier}
}

@article{silva2014toward,
  title={Toward embedded detection of polyps in wce images for early diagnosis of colorectal cancer},
  author={Silva, Juan and Histace, Aymeric and Romain, Olivier and Dray, Xavier and Granado, Bertrand},
  journal={International journal of computer assisted radiology and surgery},
  volume={9},
  number={2},
  pages={283--293},
  year={2014},
  publisher={Springer}
}

@article{vazquez2017benchmark,
  title={A benchmark for endoluminal scene segmentation of colonoscopy images},
  author={V{\'a}zquez, David and Bernal, Jorge and S{\'a}nchez, F Javier and Fern{\'a}ndez-Esparrach, Gloria and L{\'o}pez, Antonio M and Romero, Adriana and Drozdzal, Michal and Courville, Aaron},
  journal={Journal of healthcare engineering},
  volume={2017},
  number={1},
  pages={4037190},
  year={2017},
  publisher={Wiley Online Library}
}

@article{tajbakhsh2015automated,
  title={Automated polyp detection in colonoscopy videos using shape and context information},
  author={Tajbakhsh, Nima and Gurudu, Suryakanth R and Liang, Jianming},
  journal={IEEE transactions on medical imaging},
  volume={35},
  number={2},
  pages={630--644},
  year={2015},
  publisher={IEEE}
}

@article{isensee2021nnunet,
  title={nnU-Net: a self-configuring method for deep learning-based biomedical image segmentation},
  author={Isensee, Fabian and Jaeger, Paul F and Kohl, Simon AA and Petersen, Jens and Maier-Hein, Klaus H},
  journal={Nature methods},
  volume={18},
  number={2},
  pages={203--211},
  year={2021},
  publisher={Nature Publishing Group US New York}
}

@inproceedings{fan2020pranet,
  title={Pranet: Parallel reverse attention network for polyp segmentation},
  author={Fan, Deng-Ping and Ji, Ge-Peng and Zhou, Tao and Chen, Geng and Fu, Huazhu and Shen, Jianbing and Shao, Ling},
  booktitle={International conference on medical image computing and computer-assisted intervention},
  pages={263--273},
  year={2020},
  organization={Springer}
}

@inproceedings{hu2026geometric,
  title={Geometric correspondence constrained pseudo-label alignment for source-free domain adaptive fundus image segmentation},
  author={Hu, Zhouhongyuan and Zhang, Lei and Wang, Lituan and Zhang, Zhenwei and Zhu, Minjuan and Wang, Zhenbin},
  booktitle={Proceedings of the AAAI Conference on Artificial Intelligence},
  volume={40},
  number={6},
  pages={4976--4984},
  year={2026}
}

@article{wang2026eauwseg,
  title={EAUWSeg: Eliminating Annotation Uncertainty in Weakly-Supervised Medical Image Segmentation},
  author={Wang, Lituan and Zhang, Lei and Wang, Yan and Wang, Zhenbin and Zhang, Zhenwei and Yi, Zhang},
  journal={IEEE Transactions on Emerging Topics in Computational Intelligence},
  year={2026},
  publisher={IEEE}
}

@article{dong2023polyppvt,
  title={Polyp-PVT: Polyp Segmentation with Pyramid Vision Transformers},
  author={Dong, Bo and Wang, Wenhai and Fan, Deng-Ping and Li, Jinpeng and Fu, Huazhu and Shao, Ling},
  journal={CAAI Artificial Intelligence Research},
  volume={2},
  pages={9150015},
  year={2023},
}

@article{kamara2025mdpnet,
  title={MDPNet: Multiscale dynamic polyp-focus network for enhancing medical image polyp segmentation},
  author={Kamara, Alpha Alimamy and He, Shiwen and Fofanah, Abdul Joseph and Xu, Rong and Chen, Yuehan},
  journal={IEEE Transactions on Medical Imaging},
  year={2025},
  publisher={IEEE}
}

@article{huang2021hardnetmseg,
  title={Hardnet-mseg: A simple encoder-decoder polyp segmentation neural network that achieves over 0.9 mean dice and 86 fps},
  author={Huang, Chien-Hsiang and Wu, Hung-Yu and Lin, Youn-Long},
  journal={arXiv preprint arXiv:2101.07172},
  year={2021}
}

@article{zhu2024medsam2,
  title={Medical sam 2: Segment medical images as video via segment anything model 2},
  author={Zhu, Jiayuan and Hamdi, Abdullah and Qi, Yunli and Jin, Yueming and Wu, Junde},
  journal={arXiv preprint arXiv:2408.00874},
  year={2024}
}

@inproceedings{yan2025samed2,
  title={Samed-2: Selective memory enhanced medical segment anything model},
  author={Yan, Zhiling and Song, Sifan and Song, Dingjie and Li, Yiwei and Zhou, Rong and Sun, Weixiang and Chen, Zhennong and Kim, Sekeun and Ren, Hui and Liu, Tianming and others},
  booktitle={International Conference on Medical Image Computing and Computer-Assisted Intervention},
  pages={540--550},
  year={2025},
  organization={Springer}
}

@article{cheng2023sammed2d,
  title={Sam-med2d},
  author={Cheng, Junlong and Ye, Jin and Deng, Zhongying and Chen, Jianpin and Li, Tianbin and Wang, Haoyu and Su, Yanzhou and Huang, Ziyan and Chen, Jilong and Jiang, Lei and others},
  journal={arXiv preprint arXiv:2308.16184},
  year={2023}
}

@article{zhao2023fastsam,
  title={Fast segment anything},
  author={Zhao, Xu and Ding, Wenchao and An, Yongqi and Du, Yinglong and Yu, Tao and Li, Min and Tang, Ming and Wang, Jinqiao},
  journal={arXiv preprint arXiv:2306.12156},
  year={2023}
}

@inproceedings{wong2024scribbleprompt,
  title={ScribblePrompt: Fast and Flexible Interactive Segmentation for Any Biomedical Image},
  author={Wong, Hallee E and Rakic, Marianne and Guttag, John and Dalca, Adrian V},
  booktitle={European Conference on Computer Vision},
  pages={207--229},
  year={2024}
}

@article{zhao2025weakpolypsam,
  title={WeakPolyp-SAM: Segment Anything Model-driven weakly-supervised polyp segmentation},
  author={Zhao, Yiming and Zhou, Tao and Gu, Yunqi and Zhou, Yi and Zhang, Yizhe and Wu, Ye and Fu, Huazhu},
  journal={Knowledge-Based Systems},
  volume={322},
  pages={113701},
  year={2025},
  publisher={Elsevier}
}

@article{wang2025weakmedsam,
  title={Weakmedsam: Weakly-supervised medical image segmentation via sam with sub-class exploration and prompt affinity mining},
  author={Wang, Haoran and Huai, Lian and Li, Wenbin and Qi, Lei and Jiang, Xingqun and Shi, Yinghuan},
  journal={IEEE Transactions on Medical Imaging},
  year={2025},
  publisher={IEEE}
}

@inproceedings{zhang2024wesam,
  title={Improving the generalization of segmentation foundation model under distribution shift via weakly supervised adaptation},
  author={Zhang, Haojie and Su, Yongyi and Xu, Xun and Jia, Kui},
  booktitle={Proceedings of the IEEE/CVF Conference on Computer Vision and Pattern Recognition},
  pages={23385--23395},
  year={2024}
}

@inproceedings{dai2015boxsup,
  title={Boxsup: Exploiting bounding boxes to supervise convolutional networks for semantic segmentation},
  author={Dai, Jifeng and He, Kaiming and Sun, Jian},
  booktitle={Proceedings of the IEEE international conference on computer vision},
  pages={1635--1643},
  year={2015}
}

@inproceedings{zhong2024gazemedseg,
  title={Weakly-supervised medical image segmentation with gaze annotations},
  author={Zhong, Yuan and Tang, Chenhui and Yang, Yumeng and Qi, Ruoxi and Zhou, Kang and Gong, Yuqi and Heng, Pheng Ann and Hsiao, Janet H and Dou, Qi},
  booktitle={International Conference on Medical Image Computing and Computer-Assisted Intervention},
  pages={530--540},
  year={2024},
  organization={Springer}
}

@inproceedings{wu2025gnan,
  title={Graph-Based Neighbor-Aware Network for Gaze-Supervised Medical Image Segmentation},
  author={Wu, Shaoxuan and Chen, Jingkun and Jin, Zhuo and Zhang, Peilin and Gao, Zhizezhang and Feng, Jun and Zhang, Xiao and Shen, Dinggang},
  booktitle={International Conference on Medical Image Computing and Computer-Assisted Intervention},
  pages={229--239},
  year={2025},
  organization={Springer}
}

@article{wu2026gitnet,
  title={GiTNet: A graph-based trajectory-informed network for gaze-supervised medical image segmentation},
  author={Wu, Shaoxuan and Zhang, Xiao and Chen, Jingkun and Zhang, Xinyu and Xing, Yaqiong and Feng, Jun},
  journal={Medical Image Analysis},
  pages={104112},
  year={2026},
  publisher={Elsevier}
}

@article{chen2025fgi,
  title={From gaze to insight: Bridging human visual attention and vision language model explanation for weakly-supervised medical image segmentation},
  author={Chen, Jingkun and Duan, Haoran and Zhang, Xiao and Gao, Boyan and Grau, Vicente and Han, Jungong},
  journal={IEEE Transactions on Medical Imaging},
  year={2025},
  publisher={IEEE}
}

@article{ge2026gam,
  title={Adaptation follow human attention: Gaze-assisted medical segment anything model},
  author={Ge, Rongjun and Li, Ruiyi and Wang, Chong and Liu, Yuxin and Zhu, Heng and Coatrieux, Jean-Louis and Zhang, Daoqiang and Lu, Jian and Chen, Yang and Li, Shuo and others},
  journal={IEEE Transactions on Circuits and Systems for Video Technology},
  year={2025},
  publisher={IEEE}
}

@article{carion2025sam3,
  title={Sam 3: Segment anything with concepts},
  author={Carion, Nicolas and Gustafson, Laura and Hu, Yuan-Ting and Debnath, Shoubhik and Hu, Ronghang and Suris, Didac and Ryali, Chaitanya and Alwala, Kalyan Vasudev and Khedr, Haitham and Huang, Andrew and others},
  journal={arXiv preprint arXiv:2511.16719},
  year={2025}
}

\end{document}